\documentclass{article}
\usepackage{spconf,amsmath,graphicx}
\usepackage{multirow}
\usepackage{subfig}
\usepackage{caption}
\usepackage{hyperref} 

\usepackage{hyperref}

\usepackage{graphicx}
\usepackage{booktabs}
\usepackage{pgfplots}
\usepackage{amsmath}
\usepackage{float}
\usepackage{multirow}

\title{ConsPrompt: Exploiting Contrastive Samples for Few-shot Prompt Learning}
\name{Jinta Weng $^{ \dagger}$ \qquad Yifan Deng $^{ \dagger}$  \qquad Donghao Li $^{ \dagger}$ \qquad Hao You $^{ \dagger}$  \qquad Yue Hu$^{ \dagger \star}$ \qquad Heyan Huang$^{\dagger \dagger \star}$}

\address{$^{\dagger}$ Institute of Information Engineering, Chinese Academy of Sciences, Beijing, China \\
$^{\dagger}$School of Cyber Security, University of Chinese Academy of Sciences, Beijing, China \\
	$^{\dagger\dagger}$Beijing Institute of Technology, School of Computer Science and Technology, Beijing, China
	\\ $^{\dagger\dagger}$Southeast Academy of Information Technology, Putian, China
}
\begin{document}
	%
	\maketitle
\begin{abstract}
Prompt has become an effective linguistic tool for utilizing pre-trained language models. However, in few-shot scenarios, subtle changes of prompt's design always make the result widely different, and the prompt learning methods are also easy to overfit the limited samples. To alleviate this, we explore utilizing suitable contrastive samples and multi-degree contrastive learning methods to improve the robustness of prompt's representation. Therefore, the proposed Consprompt combined with prompt encoding network, contrastive sampling modules, and contrastive scoring modules, is introduced to realize differential contrastive learning. Our results exhibit the state-of-the-art performance in different few-shot settings, and the ablation experiments also certify the effectiveness of utilizing multi-degree contrastive learning in prompt-based fine-tuning process. 

\end{abstract}
\begin{keywords}
	Prompt learning, Pre-trained language model, contrastive learning, few-shot learning 
\end{keywords}
\section{Introduction}

With the exponential growth of Large language models(LLMs) and prompting strategies, pre-trained language models now could serve as a huge knowledge base or superior linguistic tool for few-shot learning \cite{radford2019language}, like ChatGPT and LLAMA. To activate the existing knowledge embedded in PLMs, researches are mainly concentrating on different prompting strategies and prompt learning methods. \cite{Madotto202110}, and zhong et al. put forward a OptiPrompt model for factual probing based on soft-prompt embedding \cite{zhong2021factual}. In terms of training strategies, prompt learning could classify into the fine-tuning of prefix-prompt and fine-tuning of independent prompt encoder, such as P-tuning\cite{liu2021gptunderstand}, Prefix-tuning\cite{li-liang-2021-prefix}, Auto-prompt learning\cite{shin2020autoprompt}, and Prompt-tuning\cite{schick2021exploiting}. 

However, subtle changes of prompt's design always make the result widely different, and the prompt learning process is also easy to over-fit the limited samples. Therefore, utilizing different cognitive learning abilities and prompting strategies to activate LLMs in few-shot settings is still a considerable question \cite{roberts2020,liu2021gptunderstand,schick2021exploiting}.
In addition to the prompting strategies, human can also use different learning rules to realize knowledge activation based on the few samples\cite{xu2023contrastive}, like contrastive learning of different prompt and different samples. Motivated by this, we exploit how to use fewer samples and multiple contrastive learning methods in prompt-based fine-tuning models.

\begin{figure}[]
 	\centering
 	\includegraphics[width=0.5\textwidth]{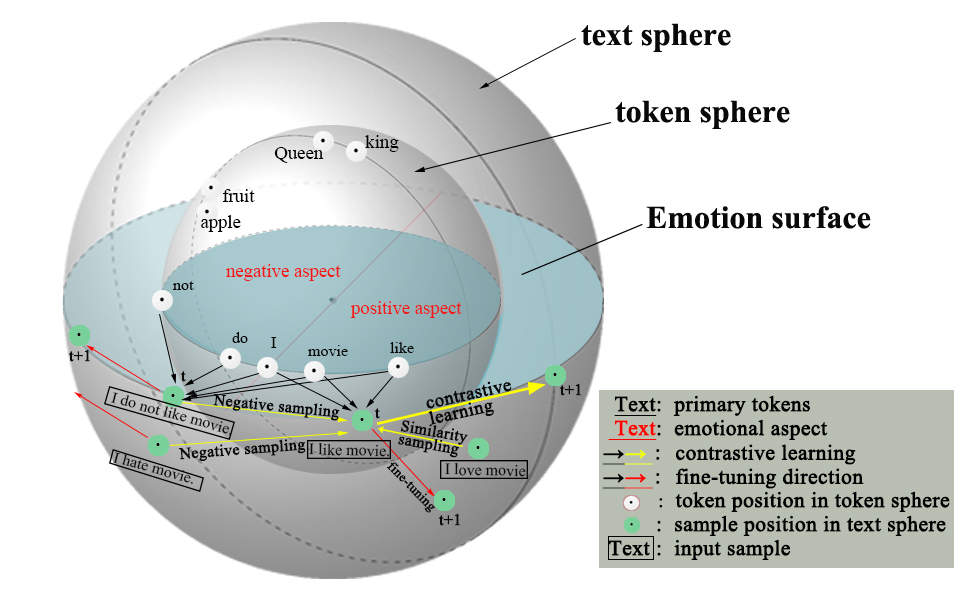}
 	\caption{The word and sentence distribution in attention-based model. We plotted contrastive learning process with the yellow row, and it reveals negative samples could contribute to the LLM's fine-tuning process and text representation step by step.}
 	\label{fig:figure2}
 \end{figure}

As shown in Fig. \ref{fig:figure2}, it reveals the token and sentence-level attention mechanism of transformer-based model, which shows the effectiveness of LLMs can be enhanced by more accurate token-level attention and contrastive learning methods. Therefore, we utilize the prompt-level and the batch-level contrastive information of tokens to strengthen few-shot prompt-based learning task, and two negative sampling strategies and SBERT \cite{reimers2019sentence} model are also used to facilitate the contrastive prompt learning model -- ConsPrompt. Our result shows its effectiveness of multiple contrastive learning in using prompt-based fine-tuning works. And the robustness of prompt-based learning can empower with suitable negative sampling strategies.
The main contribution is:




 %

\begin{itemize}
	\item We proposed a novel contrastive prompt model ConsPrompt for increasing the prompt’s distinguishing ability on prompt-level and batch-level samples.
	\item Two sampling strategies and the SimCLR contrastive learning method, are introduced to alleviate the widespread overfitting phenomenons in prompt learning process.
	\item The results demonstrate the state-of-the-art performance and show its k-shot robustness on five representative few-shot tasks.
\end{itemize}

\section{Model}
\begin{figure*}[htbp]
	\centering
	\includegraphics[width=0.65\textwidth]{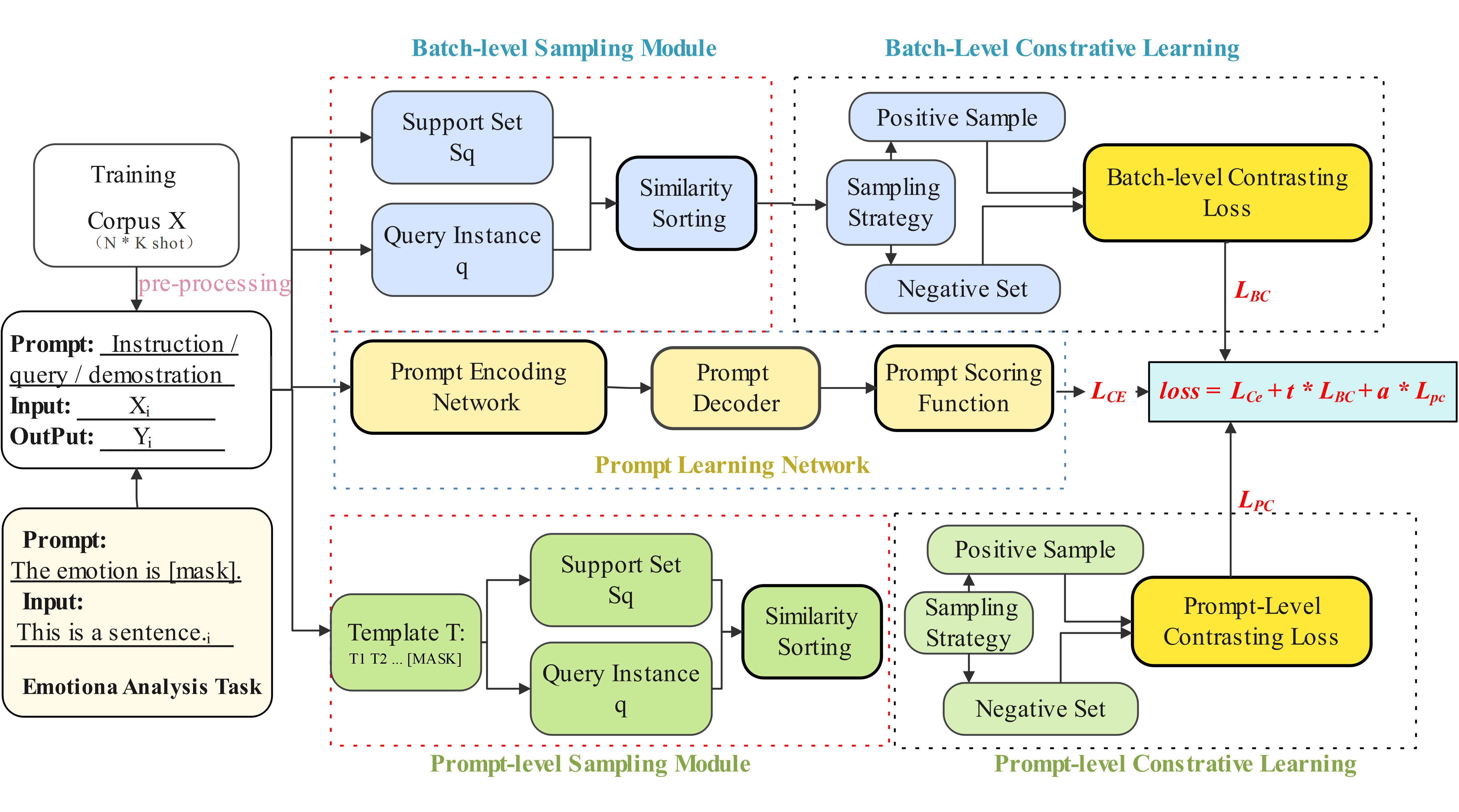}
	\caption{The model of ConsPrompt integrating prompt encording network and contrastive sampling module and scoring module} 
	\label{fig:figure3}
	\vspace{-0.5cm}
\end{figure*}

 As depicted in fig. \ref{fig:figure3}, the $\mathrm{ConsPrompt}$ are constituted by prompt-based Encoding Network, contrastive sampling module, and contrastive learning method$\mathrm{ConsPrompt} $. In prompt encoding network  (\S \ref{sec:prompt-learning}), the input would be encoded to a prompting input, and then predict the mapping tokens by prompt-based fine-tuning method. The core component of ConsPrompt is utilizing two contrast-aware prompt learning modules, which containing the prompt-level and the batch-level contrastive learning strategies. Both learning strategies containing a specif contrastive sampling module  (\S \ref{sec:Contrastive-sampling}) and learning encoder (\S \ref{sec:Contrastive-learning}. In contrastive sampling modules, the prompting input would extract negative and positive samples by similarity-based or label-based sampling methods, while these samples are subsequently used to construct the supporting set for multiple contrastive learning (\S \ref{sec:Contrastive-learning}). Finally, we use a joint loss to fine-tune the prompt encoding network and two contrastive learning encoders. 

\subsection{Prompt Encoding Network }\label{sec:prompt-learning}
We take prompt-based fine-tuning model \cite{gao2020making} as the baseline of our prompt encoding Network. 
In this method, a template $T$ consisting of textual tokens and \textit{mask} token would be defined first. Consequently, the original input $x_{i}$ would then be transformed to ${x}^{i}$. For example, if the template is adding $ It\ is\ [MASK] $ after origin input, the prompting input $\overline{x}^{i}$ would be:
\begin{equation}
	\overline{x}^{i} = T(x^{i})= x^{i}. It\ is\ [MASK]
\end{equation}
, where the hidden value of [MASK] token is used to generate the word distribution over PLMs vocabulary consequently.

Next, a label mapping is defined to map each label to specif token. The detail formulation of label mapping is defined by:
\begin{equation}	
	F(y): y_{t} \rightarrow v_{t}, y_{t} \in {Y}, v_{t} \in {V}
\end{equation}
,where $v_{t}$ is one of tokens in PLMs vocabulary, and $s$ the label index.
And the final prediction would formulate in following equation:

\begin{equation}
	\begin{split}
		p(y^{t}|x^{i}) &= p(v^{t}| 	h_{[MASK]}) \Rightarrow p(v^t |W \cdot h_{[MASK]}) \\
		& 	\Rightarrow p(v^t | h_{v_{t}}) = \ln \dfrac{\exp(h_{v_{t}})}{\begin{matrix} \sum_{k=1}^{|Y|}\exp(h_{v_{k}}) \end{matrix}}
	\end{split}
\end{equation}
where $h$ is the hidden representation of last layer in specif token, the $W$ is a vocabulary projector with the size of $|h \times V|$ , $v$ used to represent each token of the label mapping.\par
The fine-tuning loss of prompt encoding network is:
\begin{equation}
	L^{CE}=-\dfrac{1}{N}\sum_i\sum_{t=1}^{|Y|} y_{i}\log{p(y^{t}|	\overline{x}^{i})}
\end{equation}
where $i$ is the index of the training pair $(x_{i},y_{i})$ and |Y| is the total number of labels.
\subsection{Contrastive Sampling Module}
\label{sec:Contrastive-sampling}
The essence of contrastive learning is to learn the differences between limited positive and negative samples. Therefore, a suitable sampling strategies for selecting positive and negative samples is essential to develop contrastive learning.\par
In the sentence representation of contrastive learning, different prompts(similar as the context knowledge) always cause different sentence representation, and different comparative objects also result in differential comparative loss. 
We thus use two sampling strategy to construct the support set $ \textit{Sq} $: prompt-level sampling and batch-level sampling module. In prompt-level sampling, we use different prompt template to construct different querying inputs for constructing the supporting set. In batch-level sampling, we use different samples of current batch to construct the support set. 

We then calculate the cosine similarity between the representation of current instance $ q $ and each supporting set $ \textit{Sq} $ to choose the positive and negative samples. The detailed formulation is:
\begin{gather} 
	MaxRank(sim(M_{sbert}(q),M_{sbert}(S_{q})))
\end{gather}
where n is the number of supporting set $S_{q}$ used to generate contrastive samples, and $q$ is the query sample. The final support set would be ordered by the max similarity. 
\par
After generating the similarity-based order in $S_{q}$ , the representation of the positive samples and negative samples still use the same PLM of prompt encoding network.  
\begin{gather} 
s^{n}_{q}= PLM(S^{n}_{q})
\end{gather}
where $ s^{n}_{q} $ is the PLM's representation of $ S^{n}_{q} $. 

What is more, we only use one positive sample selecting by the highest similarity of support set, while negative samples are selected by the lower similarity among in other samples with different labels.
\subsection{Contrastive Scoring Module}
\label{sec:Contrastive-learning}
Contrastive scoring modules are core component using to learn the representation of sample distinction. After above-mentioned sampling operation, we use same PLM of prompt encoding network to representing each samples.
\begin{gather}
	S_{BC,PC}(i,j) = -\log\frac{\exp(s_{i,j}/\tau )}{ {\textstyle \sum_{k}^{N_{neg}}1_{[k\ne i]}\exp(s_{i,k}/\tau)} }
\end{gather}
where \textit{k} is the index of negative sample set, $\tau$ is a controllable temperature parameter, S$_{BC}$ and S$_{PC}$ is the batch-level and prompt-level supporting sets. \par
We utilize the SimCLR  \cite{chen2020simple} to compute two comparative losses, the S$_{BC}$ and the S$_{PC}$ of prompt-level and batch-level sampling module. Owing to the difference negative support set of the comparative objects, we also switch their position in $S(i,j)$ to calculate another inverted loss. The final loss of contrastive learning modules are thus defined as:
\begin{gather}
		\begin{split}
	L_{BC} =\frac{1}{{N}}  {\textstyle \sum_{N}^{}}\left [S_{BC}(i,j) + S_{BC}(j,i)   \right ]\\
	L_{PC} =\frac{1}{{N}}  {\textstyle \sum_{N}^{}}\left [S_{BC}(i,j) + S_{PC}(j,i)   \right ]
	\end{split}
\end{gather} 
where N is the samples number of prompt-level or batch-level supporting sets.
Considering the loss of contrastive ability and prompt mechanism, the final loss is formulated as:
\begin{gather}
	L =  L_{CE} + t\ast L_{BC} + a \ast L_{PC} 
\end{gather} 
where \textit{t} and \textit{a} is the hyper-parameter to determine the ratio of comparative learning loss.


\begin{table*}[htbp]
	\centering
	\scalebox{0.75}{
	\begin{tabular}{lccccc}
		\toprule
		\textbf{Baselines} \  &\textbf{ TREC (acc)} & \textbf{SNIL (acc)} &\textbf{ QNLI (acc)} & \textbf{SST-5 (acc)}& \textbf{ SST-2 (acc) } \\
		\midrule
		Majority                            & 18.8      & 33.8      & 49.5 & 50.9      & 23.1      \\
		prompt-based zero-shot learning     & 32.0      & 49.5      & 50.8        & 35.0   & 83.6    \\
		GPT3-in-context-learning            & 26.2(2.4) & 47.1(0.6) & 53.8(0.4)    & 30.6(0.9) & 84.8    \\
		fine-tuning                         & 27.2(1.4) & 48.4(4.8) & 60.2(6.5)  & 43.9(2.0) & 81.4(3.8) \\
		\midrule
		Prefix-tuning$  $ & 36.0(1.1) & 33.5(3.9) & 54.5(2.2) & 46.1(1.3)  & 88.1(2.3) \\
		P-tuning$  $ & 40.2(1.3) & 37.5(1.6) & 57.6(3.9) & 32.1(3.1)  & 90.1(1.2) \\
		LMBFF & 84.8(5.1) & 77.1(3.9) & 64.5(4.2) & 46.1(1.3)  & 92.1(1.1) \\ \midrule
		CP-Tuning & -- & --& 69.22 & -- & 93.35\\
		Jian et al., 2022 &83.3 (1.5)& 69.9 (2.4) &66.4 (3.5) &\textbf{ 49.5 (1.1)} & 90.6 (0.1)\\
		\midrule
		ConsPrompt(-sim)    &\textbf{ 87.5 (2.2)} & \textbf{77.3 (3.6)} &\textbf{72.0 (3.0)}  &   47.2 (3.1) &   \textbf{95.0(2.8)} \\
		ConsPrompt(-label)    &  86.8 (2.8)  &   76.2 (3.8)  &71.9 (2.7)  &   47.5 (2.4) &    93.1 (1.4)  \\
		\bottomrule
		
	\end{tabular}
}
	\caption{The main result of The ConsPrompt. We select the baselines from prompt learning modules(prefix-tuning, p-tuning), LMBFF model\cite{gao2020making}), and two contrastive prompt framework, the CP-tuning(\cite{xu2022making} and the Jiang's \cite{jian-etal-2022-contrastive} work.}
	\label{tab:tabel1}
		\vspace{-0.5cm}
\end{table*}
	\begin{table}[]
	\centering
		\scalebox{0.8}{
	\begin{tabular}{clll}
		\hline
		\multirow{2}{*}{\textbf{ t,a }} & Average & Variance & Median \\ 
		& (acc.)&(+std)&(acc.) \\ \hline
		0.1                  & 75.5          & 2.7      & 76.0   \\
		0.5                  & 75.7          & 2.4      & 76.0   \\
		1.0                  & 75.5          & 3.2      & 76.1   \\
		20                   &\textbf{ 77.3  }        & 3.6      & \textbf{78.9}   \\ \hline
		\textbf{Ensemble}    & 77.0          & 2.9      & 77.7   \\ \hline
	\end{tabular}
}
	\caption{The comparative experiment using different ratios of comparative scoring module. We choose ConsPrompt(-sim) as the baseline and change the t and a on 0.1, 0.5, 1.0 and 20.}
	\label{Tab:tab2}
		\vspace{-0.5cm}
\end{table}

\section{Experiment}
In this section, we introduce our using datasets, experimental setup, the main results of our experiments and its analyses.
\subsection{Few-shot Experiment setups}
We use RoBERTa-large as our pre-training language model. Our experiment are developed in NVIDIA V100 32GB (also could run in 1080ti with low batch size). In the training process, we develop many experiments on different batch sizes \textit{bs=4,8,16}, learning rate \textit{lr={1e-5,2e-5,5e-5}}. \par
We evaluate our proposed model ConsPrompt on few-shot glue tasks, including question-answering task (TREC dataset), emotion classification tasks (the sst-2 and sst-5 datasets), and text entailment tasks(QNLI and SNLI datasets). To satisfy the few-shot learning setting, we pick five different K-shot sub-datasets and each sub-dataset is constructed by K=16 training pairs on each type of labels\cite{gao2020making}. Also, we use the mean score and variance of the prediction result on different subsets instead of the highest result to control the fairness.
In the setting of contrastive learning, we select all instances except query instance q to combine the initial supporting set. In order to decree the calculation within initial supporting samples, we set the filtering ratio of the support set to 0.5. We set $\tau$ = 0.07 to realize the smoothing of loss, and the ratio $t$ and $a$ of contrastive learning are 0.5. Our source is available at https://github.com/Nagin-Kim/cosprompt.  
\vspace{-0.5cm}
\subsection{Baselines}
We compare with the Majority method (select the majority class as prediction), fine-tuning method\cite{Liu2019RoBERTaAR}, prompt-based zero-shot learning, GPT3-in-context-learning\cite{Brown2020Language}, prompt learning modules(prefix-tuning, p-tuning), LMBFF model\cite{gao2020making}), and two contrastive prompt framework, the CP-tuning(\cite{xu2022making}, only certificate in binary classification tasks) and \cite{jian-etal-2022-contrastive}(label-based sampling method).
Based on different sampling strategies, our ConsPrompt is divide into two  label-based and sim-based ConsPrompt. In ConsPrompt(-sim), the negative and positive samples are directly sampled based on the similarity between the support sets and query instance, while ConsPrompt(-label) are label-based sampling strategy.
	\vspace{-0.5cm}
\subsection{Main Result}
The main result of ConsPrompt is depicted in Tab. \ref{tab:tabel1}. Comparing with existing baselines, the ConsPrompt model achieves the state-of-the-art results. It reveals that the prompt learning process indeed increases the model's discrimination on negative and positive samples. What is more, the effectiveness of ConsPrompt(-sim) is better than ConsPrompt(-label) in TREC, SNLI and SST tasks. Since the sampling samples from ConsPrompt(-sim) use a similarity-first viewpoint on generating negative set from SBERT, it seems that we need to integrate more information about the fine-grained semantic distinction for few-shot prompt learning, instead of the coarse label distinction. In addition, the variances result of the ConsPrompt are generally lower than other baselines, which shows its greater robustness to some extent.  
\begin{table}[]
	\scalebox{0.75}{
	\centering
	\begin{tabular}{lclllc}
		\hline
		\textbf{Tasks}  &\textbf{ Fine-tuning} & \multicolumn{2}{c}{\textbf{Our Sim-based }} & \multicolumn{2}{c}{\textbf{Our Label-based}} \\ 
		K-shot         & Acc        & Acc(+std)         & Mean.        & Acc (+std)       & Mean.      \\\hline
		8                        & 16.5  & \textbf{47.1 (1.6)}            & 47.3           & 44.5 (2.1)       & 44.4        \\
		16                 &  19.8       & 47.2 (3.1)            & 46.9           & \textbf{47.5 (2.4)}       & 46.5         \\
		32                   & 22.4     & 48.8 (1.6)            & 48.9           & \textbf{48.3 (1.3) }       & 48.6        \\
		64                & 28.2      &\textbf{ 51.1 (1.0) }           & 50.9           & 50.8 (0.9)       & 50.6         \\
		128             &35.9        & \textbf{51.8 (2.0) }           & 52.4           & 51.5 (1.1)       & 51.4         \\
		160          & 42.1            & \textbf{51.8 (1.3)    }        & 51.9           & 51.7 (0.8)       & 51.8         \\ \hline
	\end{tabular}
}
	\caption{The result using different K sampling strategies. K is the samples of per classes in current datasets.}
	\label{Tab:tab3}
	\vspace{-0.4cm}
\end{table}

\subsection{Analysis}
We explore different temperatures and K-shot settings to certificate the Consprompt's necessity and effectiveness.
\subsubsection{Different Ratio of Contrastive Loss.}
In order to explore the effectiveness of the contrastive learning module, we choose SNLI task and set different ratio of t and \textit{a} to control the loss from contrastive scoring module.
As the result shown in Tab. \ref{Tab:tab2}, with the increase of value \textit{t} in comparative learning module, the proposed ConsPrompt is able to receive more gain from the contrastive learning module, since it does not make any big gains while setting the lower t values, and higher t value can empower the original prompt encoding network. In addition, the integration result of different t settings reveals the necessity of comparative learning module.
\subsubsection{The K-shot Robustness of ConsPrompt.}
We also consider the influence of different sampling numbers in few-shot datasets. We set the value K to 8,16,32,64,128,160 on SST-5 tasks. 
As the result shown in Tab. \ref{Tab:tab3}, the higher K-shot setting can bring more effectiveness on the accuracy and robustness. The ConsPrompt (Sim-based) are effective on higher K-setting experiments, while label-based ConsPrompt is effective in 16-shot settings. It reveals the similarity-based sampling strategy is more efficient than label-based strategy. Also, with the increase of K, the benefit from the training corpus is decreasing, which shows the consprompt are more suggestive for few-shot experiments.
%
%
\vspace{-0.5cm}
\section{Conclusion}

Combining prompt learning and unsupervised contrastive learning, we propose an efficient prompt model ConsPrompt for few-shot scenario. The ConsPrompt integrating prompting encoding network, Contrastive sampling module and contrastive scoring module, is able to realize multiple learning and alleviate the over-fit problem in prompt design .The effectiveness of ConsPrompt on few-shot  learning tasks (only 16 samples per class) shows state-of-the-art performances and more sample-degree robustness. Our future work will focus on improving the controllability of comparison sample selection.
\par
\
\section{Acknowledgment}
This work is supported by the National Natural Science Foundation of China (Grant No.U21B2009) and the Science Foundation of (Grant No.E110151101, E250471101).
\bibliographystyle{IEEE}
\bibliography{ref}

\end{document}